# DeepAISE - An End-to-End Development and Deployment of a Recurrent Neural Survival Model for Early Prediction of Sepsis

Supreeth P. Shashikumar[1,4], Christopher Josef[2], Ashish Sharma[3], Shamim Nemati[4]*

**Affiliations:**

[1]School of Electrical and Computer Engineering, Georgia Institute of Technology, Atlanta, USA.

[2]Department of Surgery, Emory University School of Medicine, Atlanta, USA.

[3]Department of Biomedical Informatics, Emory University School of Medicine, Atlanta, USA.

[4]Department of Biomedical Informatics, University of California San Diego Health, La Jolla, USA.

*To whom correspondence should be addressed: shamim.nemati@alum.mit.edu

**Abstract:** Sepsis, a dysregulated immune system response to infection, is among the leading causes of morbidity, mortality, and cost overruns in the Intensive Care Unit (ICU). Early prediction of sepsis can improve situational awareness amongst clinicians and facilitate timely, protective interventions. While the application of predictive analytics in ICU patients has shown early promising results, much of the work has been encumbered by high false-alarm rates. Efforts to improve specificity have been limited by several factors, most notably the difficulty of labeling sepsis onset time and the low prevalence of septic-events in the ICU. Here, we present DeepAISE (Deep Artificial Intelligence Sepsis Expert), a recurrent neural survival model for the early prediction of sepsis. We show that by coupling a clinical criterion for defining sepsis onset time with a treatment policy (e.g., initiation of antibiotics within one hour of meeting the criterion), one may rank the relative utility of various criteria through offline policy evaluation. Given the optimal criterion, DeepAISE automatically learns predictive features related to higher-order interactions and temporal patterns among clinical risk factors that maximize the data likelihood of observed time to septic events. DeepAISE has been incorporated into a clinical workflow, which provides real-time hourly sepsis risk scores. A comparative study of four baseline models indicates that DeepAISE produces the most accurate predictions (AUC=0.90 and 0.87) and the lowest false alarm rates (FAR=0.20 and 0.26) in two separate cohorts (internal and external, respectively), while simultaneously producing interpretable representations of the clinical time series and risk factors.

**Introduction**

Sepsis is a syndromic, life-threatening condition that arises when the body's response to infection injures its own internal organs *(1)*. Though the condition lacks the same public notoriety as other conditions like heart attacks, 6% of all hospitalized patients in the United States carry a primary diagnosis of sepsis as compared to 2.5% for the latter*(2)*. When all hospital deaths are ultimately considered, nearly 35% are attributable to sepsis (2). This condition stands in stark contrast to heart attacks which have a mortality rate of 2.7-9.6% and only cost the US $12.1 billion annually, roughly half of the cost of sepsis *(3)*.

Starting in 2004 the Surviving Sepsis Campaign (SSC) began addressing the variations in clinical treatment regimens for sepsis and septic shock through the promulgation of evidence based practice guidelines called "sepsis bundles"*(4)*. These bundles consolidate the results of numerous investigations that have repeatedly demonstrated improvement in sepsis outcomes after the timely administration of broad-spectrum antibiotics, Intravenous (IV) fluids, and vasopressors when indicated *(5–8)*. The most recent recommendation from the SSC is a 1-hr bundle that in addition to obtaining diagnostic tests like cultures and lactate levels, prescribes standard treatment with broad spectrum antibiotics, IV fluid, and vasoactive drugs if necessary, all within an hour of a sepsis diagnosis *(9)*. While there are effective protocols for treating sepsis once it has been diagnosed there still exists challenges in reliably identifying septic patients early in their course.

In recent years, the increased adoption of electronic medical records (EHRs) in hospitals *(10)* has motivated the development of machine learning based surveillance tools for detection or classification *(11–15)* and prediction *(11, 16–18)* of patients with sepsis or septic shock. For prediction of sepsis in particular, Desautels et al. *(11)* used a proprietary machine learning algorithm called InSight to achieve an Area Under the Curve (AUC) of 0.78 in predicting onset of sepsis four hours in advance. Nemati et al. *(16)* used a modified Weibull-Cox model on a combination of low-resolution Electronic Medical Record (EHR) data and high-resolution vital signs time series data to predict onset of sepsis four hours in advance with an AUC of 0.85. Other works *(17, 18)* have focused on developing models to predict septic shock, which occurs when sepsis leads to low blood pressure that persists despite treatment with intravenous fluids. However, a direct comparison of these methods is not possible for several reasons: 1) utilization of different labels for sepsis and septic shock, 2) variations in prediction horizon (finite horizon prediction vs infinite horizon prediction, 3) differences in frequency of prediction (single event classification vs sequential prediction), and 4) variations in study design and disease prevalence (case-control design vs calibrated real-world prevalence models). To date most sepsis prediction research has failed to make the transition into viable Clinical Decision Support (CDS) systems owing to the relatively low clinical tolerance for false-alarms *(19)*, as well as the interpretability and workflow integration requirements for CDS systems *(20, 21)*. False clinical alarms not only increase the cognitive load on clinicians but can also expose patients to unnecessary antibiotics and may contribute to emergence of antibiotic resistance pathogens *(22)*. Nevertheless, identifying and treating true cases of sepsis before they are clinically apparent is categorically one of the most important needs for modern medicine to address.

Consistently identifying the onset time for sepsis presents unique challenges because the condition manifests as constellation of signs and symptoms with significant variability in presentation and timing. The Third International Consensus Definitions for Sepsis (Sepsis-3) guidelines have provided two primary criteria for making a formal diagnosis of sepsis: 1) There must be a suspicion for infection (indicated by the administration of antibiotics for at least 72hrs with the concomitant collection of cultures) 2) There must be a two-point increase in the SOFA (Sequential Organ Failure Assessment) score *(1)*. These criteria have associated time points and from these time points, sepsis can be consistently labeled. While the Sepsis-3 criterion is considered the current standard for labeling sepsis onset time, previous consensus criteria for sepsis (based on Sepsis-1 and Sepsis-2 definitions) *(23, 24)* remain in wide use. Additionally, there are other sepsis criteria developed by the Center for Disease Control (CDC) and Center for Medicare and Medicaid Services (CMS) for use in surveillance studies *(25, 26)*.

When applied to the same patient population, the execution of aforementioned criteria often leads to significant variability in the sepsis onset times. Therefore, a question of interest is what consensus criteria for sepsis (on average) is likely to result in improved patient outcomes if coupled with appropriate treatment protocols. In this work, we show that by coupling a criterion with a treatment policy (e.g., initiation of antibiotics within an hour of meeting the criterion), one may use the framework of Attributable treatment effect estimates (ATEE) *(27)* to rank the relative utility of various criteria through policy evaluation. The ATEE framework allows one to ask outcome-specific counterfactual questions such as 'what-if a patient had received antibiotics 4-6 hours prior to meeting a given sepsis onset time?' or 'what-if a treatment bundle requirement was met within 3-hours of clinical recognition?'. This is equivalent to evaluating the expected reward (e.g. hospital survival or 60 days survival) associated with a counterfactual treatment policy (e.g., early detection of sepsis via an automated algorithm followed by administration of antibiotics) compared to the actual hospital policy (e.g., suspicion of sepsis by a clinician followed by administration of antibiotics).

The three primary contributions of this work are:

a) An ATEE framework to compare the various consensus criteria of sepsis and choose the most optimal amongst them.
b) A deep learning framework for prediction of sepsis (called DeepAISE) that reduces incidents of false alarms by automatically learning predictive features related to higher-order interactions and temporal patterns among clinical risk factors for sepsis. Unlike comparable models, this algorithm maintains interpretability by tracking the top relevant features contributing to the sepsis score as a function of time, providing clinicians with rationale for alerts. Most importantly, DeepAISE is a generalizable model developed using 25,000 patient admissions to the Intensive Care Units (ICUs) at two Emory University hospitals and over 40,000 ICU admissions from the Medical Information Mart for Intensive Care-III (MIMIC-III) ICU database.
c) A software platform that integrates the DeepAISE framework with live streaming patient data, computes sepsis risk scores in real-time, and presents them in an interpretable, clinically meaningful fashion. The platform is scalable and can be integrated into clinical workflow.

**Results**

*Study design and patient cohorts*

The DeepAISE model was trained and validated on two separate patient cohorts. The Emory cohort consisted of all patients admitted to the ICUs at two hospitals within the Emory Healthcare system in Atlanta, Georgia from 2014 to 2018. External validation of the DeepAISE model was performed on the MIMIC-III database *(28)*, which is a publicly available database containing de-identified clinical data of patients admitted to a medical center in Boston, Massachusetts from 2001 to 2012. The complete set of patient features (Appendix C of Supplementary Material) was grouped into three categories: clinical features (e.g. heart rate, mean arterial pressure, etc.), laboratory test results (e.g. hemoglobin, creatinine, etc.) and demographic/history/context features (e.g. age, care unit type, etc.).

In both the cohorts, the Third International Consensus Definition of Sepsis (Sepsis-3) *(1, 29)* criterion was used to assign sepsis onset time ($t_{sepsis-3}$) when two conditions were simultaneously satisfied: 1) there was a clinical suspicion of infection ($t_{suspicion}$) and 2) there was a two point increase in SOFA score ($t_{SOFA}$). Similarly, all patients were labeled for the Center for Disease Control (CDC) onset time of sepsis ($t_{sepsis-CDC}$) according to a modified version of the criterion developed the CDC *(25)* when two conditions were simultaneously satisfied: 1) there was a clinical suspicion of infection and 2) there was an eSOFA score of one or higher ($t_{eSOFA}$) *(30)*. The labels of $t_{suspicion}$, $t_{SOFA}$, $t_{sepsis-3}$, $t_{eSOFA}$, and $t_{sepsis-CDC}$ are used extensively throughout this study to define key time points and are clearly described in Table 1.

The Emory cohort consisted of a total of 25,820 patients (see Table S1 for patient characteristics), 1,445 of whom met the Sepsis-3 criterion four hours or later after ICU admission *(29)*. Those who developed sepsis tended to have a slightly higher percentage of male patients compared to non-septic patients (55.2% vs. 53.2%) and had more comorbidities (Charlson Comorbidity Index [CCI] 3 vs. 2). Septic patients had longer median lengths of ICU stay (5.9 vs. 1.9 days), higher median SOFA scores (5.0 vs. 1.7), and higher hospital mortality (15.2% vs. 3.5%). The median [interquartile range (IQR)] time from ICU admission to $t_{sepsis-3}$ in the Emory cohort was 24 [9, 63] hours. In comparison, the median [IQR] time from ICU admission to $t_{sepsis-CDC}$ in the Emory cohort was 31 [11, 70] hours. Similar patterns were observed for the MIMIC-III external validation cohort (Appendix E of Supplementary Material, Table S3).

*Attributable treatment effect estimates*

We used the framework of attributable treatment effect-size estimates (ATEE) to quantitatively compare the two aforementioned consensus criteria (namely $t_{sepsis-3}$ and $t_{sepsis-CDC}$) for labeling sepsis onset time (see Material and Methods for more details). We considered four different treatment levels corresponding to different antibiotic administration delay intervals with respect to the onset time of sepsis, namely [-24, –6), [-6, 0), [0, 3), and [3, 24] hours. The ATEE framework was used to assess the survival effects of early or late antibiotics administration with respect to each criterion (see Table 2). Entries in the table correspond to expected improvement in outcome (defined as not dying or not getting transferred to a hospice) as estimated by Eq. 1. In all cases, early administration of antibiotics resulted in a net improvement in outcome, compared to the existing clinical policy. The results from Table 2 suggest that if a clinical policy based on prediction of $t_{sepsis-3}$ six hours in advance followed by administration of antibiotics was put in place, one could have expected 8.2 [7.4, 9.9] % improvement in patient outcomes. Whereas a clinical policy based on prediction of $t_{sepsis\text{-}CDC}$ six hours in advance followed by administration of antibiotics could have resulted in 0.8 [-0.2, 5.3] % improvement in patient outcomes.

*DeepAISE prediction performance for sepsis onset*

The DeepAISE model was trained to predict the early onset of sepsis ($t_{sepsis\text{-}3}$). DeepAISE made hourly predictions, starting four hours after ICU admission, and considered a total of 65 features that were commonly available in the EHR. The Emory training and testing sets contained a total of approximately 500,000 and 125,000 hourly prediction windows, respectively. A complete list of all the input features is provided in Appendix C of Supplementary Material.

The DeepAISE model reliably predicted $t_{sepsis\text{-}3}$ four hours in advance with an AUC of 0.90 (specificity of 0.80 at sensitivity of 0.85) on the Emory testing set. Slightly lower performance was observed for the MIMIC testing set, with the DeepAISE model achieving an AUC of 0.87 for predicting $t_{sepsis\text{-}3}$ four hours in advance (see Table 3 for more details). To assess the impact of changes in institutional practices and patient populations over time we performed an experiment in which a model trained on Emory year-based training set (patients in Emory cohort from the year 2014 through 2017) was applied to a holdout test set collected from 2017 to 2018 (Emory year-based holdout set). The DeepAISE model achieved an AUC of 0.88 on the Emory year-based holdout set (see Fig. S4 for more details) suggesting that the model was not adversely impacted by changes in patient population or clinical practices over time.

*DeepAISE performance against other baselines*

The DeepAISE model is a composite model made of a class of Recurrent Neural Network (RNN) models known as Gated Recurrent Units (GRUs) *(31)* that feeds into a Weibull Cox Proportional Hazards (WCPH) model (see section *'Development of the DeepAISE model'* in methods for more information). This architecture was chosen in the context of predicting sepsis onset time as a time-to-event analysis and considering that temporal changes in patients' physiology are important for the early prediction of sepsis. We assessed the utility of this model architecture by comparing performance of DeepAISE against three different baseline models: 1) a Logistic Regression (LR) model, 2) a Weibull-Cox proportional hazards (WCPH) model, and 3) a Feedforward Neural Network (FFNN) with two layers of 100 hidden units and a final WCPH layer for prediction of onset of sepsis (See Fig.1). A more comprehensive comparison of DeepAISE with other baseline models can be found in Supplementary Material (See Fig. S3).

Across all prediction windows, DeepAISE consistently outperformed all other classifiers ($p<0.001$; when AUC of DeepAISE was compared with AUC of other baseline methods) for prediction of $t_{sepsis-3}$ (See Fig. 1, Fig. S3 and Table S2) and across all prediction windows, indicating that capturing temporal trends and interactions among the risk factors is important for accurate prediction of sepsis. The performance of all the models decreased with the increase in prediction horizon. For DeepAISE, the AUC on the Emory testing set decreased from 0.90 at 4-hour prediction horizon to 0.88 at 12-hour prediction horizon. We also observed that these findings were consistent with the MIMIC testing set (Fig. S8).

In addition, a FFNN trained to predict $t_{sepsis-3}$ with delta change in SOFA score as input achieved 0.54 AUC on Emory testing set, and a FFNN trained to predict $t_{sepsis-3}$ with delta change in SOFA score and static covariates (such as age, gender, weight etc.) as inputs achieved 0.68 AUC on the Emory testing set. The above results show that DeepAISE scores were not simply recapitulations of the SOFA scores.

*Clinical interpretation of DeepAISE predictions*

While performance characteristics of machine learning algorithms are important, providing interpretable data to the bedside clinicians that can guide diagnosis and therapeutic interventions is a critical requirement of CDS systems. To date many sepsis models have failed to demonstrate which physiologic aberrations contributed to the model's prediction, compelling many to refer to them as "black boxes" DeepAISE allayed these concerns by continually revealing the top patient features contributing to its predicted risk scores. Unlike many other algorithms, DeepAISE is uniquely interpretable as evidenced in Fig. 2 in which the trajectory of a septic patient who developed pneumonia in the ICU is displayed. In this visualization, the sepsis risk score predicted by the model is shown along with vital sign trends, and most notably, the most relevant features contributing to the risk score. In this example, early deterioration of the patient's respiratory status was detected by the model. The model identified aberrations in $PaO_2$, $PaCO_2$, blood pH and Glasgow coma score (GCS) as some of the top features relevant to its prediction. The importance of each feature was calculated using the magnitude and sign of the associated relevance scores, in a fashion similar to saliency maps for convolutional neural networks *(32)*. To validate the clinical interpretability of the DeepAISE model, analysis of the most relevant features starting 10 hours prior to and ending at $t_{sepsis-3}$ was conducted (see Fig. 4). This investigation revealed that DeepAISE ascribed importance to several features that have already been identified as risk factors for sepsis such as recent surgery, length of ICU stay, heart rate, GCS, white blood cell count, and temperature, and some less appreciated but known factors such as Phosphorus (or hypophosphatemia) *(33)*. This analysis provides a global view of model interpretability, whereas the individual relevance scores provide a local view of interpretability by listing the top features contributing to the risk scores at each hourly point in time. Perturbation analysis revealed that the globally important features may not provide an accurate view of the top contributing factors to the individual risk scores. We observed that treating the top locally important features as missing values yielded a significantly lower AUC compared with a similar analysis replacing the globally most important features. (See Fig. 4; Additionally, refer to Appendix F of Supplementary Material for a more thorough analysis of global model interpretability vs local model interpretability).

*Inferring significance of individual patient trajectories*

Clinicians have long appreciated that trends in patient metrics are often more telling than discrete point values. The high dimensional nature of the data used to represent a patient is challenging to represent. Display of patient trajectories as they pass from states of sickness to health provides yet another opportunity to inform the clinician about a patient's expected clinical course.

Each point on the manifold shown in Fig. 3 is a 3D representation (projection) of the patient's 65 features, constructed via first learning a 100-dimensional representation (last layer of the DeepAISE model) followed by dimensionality reduction via Spectral clustering *(34)*. Two exemplar patient trajectories are presented in Fig. 3. Patient 2 was in a state of good health (specifically no suspicion for infection) prior to developing a subdural hemorrhage which prompted admission. This patient went on to be diagnosed with a ventilator associated pneumonia two days after an emergent craniectomy. In contrast Patient 1, who was several weeks status post craniectomy for stroke, was readmitted with a culture positive pneumonia present on admission. The manifold in Fig. 3 shows that trajectories for patient 1 and 2 follow similar terminal patterns; however, correctly assigns them different starting positions with patient 2 starting from a comparatively higher risk cluster. The specific trajectory of an ICU patient may be useful in categorizing infectious phenotypes and detecting anomalous physiological dynamics.

*DeepAISE user interface and tele-ICU workflow integration*

The goal of any CDS system is to improve patient outcomes and reduce hospitalization costs; however, actualization of these goals is incumbent upon clinical teams embracing and actually employing the technology. The integration of a CDS system into clinical workflows depends on many factors, and therefore the development of the DeepAISE UI (User Interface) involved nursing stakeholders in our tele-ICU center.

Appreciating the workflow of the tele-ICU staff was a critical component of ensuring that the developed UI was both useable and interpretable. Nursing stakeholders identified the key tasks in the tele-ICU as consisting of the following: routine patient assessment, sepsis risk assessment, communication with the bedside clinical team, and physician initiation of therapeutic interventions. A minimal user interface (UI) that enhanced workflow awareness, provided easy actionability, and ensured data integrity was built after soliciting requirements from the aforementioned stakeholders in early 2017.

The resultant UI shown in Fig. 6 was designed to present a list of patients sorted by sepsis risk score. Square cards that include the sepsis risk score as well as the change in score over the past hour are used to represent a single patient. Individual cards can be flipped via a single mouse-click to reveal the top factors contributing to the presented score. To improve individual and unit situational awareness regarding patient interventions and assessments, users can drag-and-drop patient cards into columns representing different treatment categories.

Data integrity assessment was carried out via manual chart-review of patient records (vitals, labs, etc.) to ensure that the displayed patient data on the user interface matched the corresponding values within the clinician facing EHR application (Cerner's PowerChart). Overall, 100% of users

(five out of five tele-ICU nurses) reported improvement in usability with the addition of top contributing factors to the risk score and workflow related features into the user interface.

**Discussion**

The major contribution of this effort is a framework for joint optimization of features (via deep learning) and labels (through counterfactual reasoning) for training predictive models in clinical settings. We showed that the ATEE framework can be used to optimally select among multiple criteria of sepsis (i.e., Sepsis-3 criterion and Sepsis-CDC criterion) while simultaneously providing an estimate of the effect size associated with early or late antibiotics administration. Moreover, a deep learning model was used to automatically learn complex features, including temporal trends and higher-order interactions among the risk factors, to accurately predict the likelihood of sepsis in patients admitted to the ICU up to 12 hours in advance. Finally, satisfaction amongst users of our CDS system was most greatly impacted by the enhancement of clinical interpretability of the findings through a workflow-aware UI that incorporated a patient's trajectory and the key factors contributing to their risk score.

DeepAISE was developed to predict two time-points of interest in this study, namely $t_{sepsis-3}$ (Sepsis-3 criterion), and $t_{sepsis-CDC}$ (Sepsis-CDC criterion). Counterfactual reasoning indicated that there would be an approximate 8% improvement in patient survival if a clinical policy of administrating antibiotics 6-hours prior to $t_{sepsis-3}$ was implemented. Note, this improvement was with respect to the actual antibiotic administration times (i.e. the clinical policy) at a tertiary academic medical center. Similar analysis revealed a marginal expected improvement in survival when considering a clinical policy based on administration of antibiotics 6-hours prior to $t_{sepsis-CDC}$. Notably, the onset time of sepsis according to the CDC criterion often occurred approximately 6-

7 hours after the sepsis-3 onset time. This delay was likely attributable to the more stringent eSOFA scoring method that requires an adverse change of $\geq 50\%$ from a patients' baseline value in order to mark an increase in individual organ scores. When considering interval of [-24, -6) hours prior to onset time of sepsis we did not observe any statistically significant difference between the expected outcomes (roughly 8% for both criteria). However, prediction of onset time of sepsis degrades with longer prediction windows and one may expect higher false alarms and lower predictive value over the [-24, -6) hours interval. We observed that the six hours ahead prediction AUC of DeepAISE (on Emory testing set) was 0.89 for $t_{sepsis-3}$. Additionally, the performance of DeepAISE expectedly dropped as the prediction window increased from 2 to 12 hours. All the findings in patients were externally validated with the MIMIC-III cohort.

Another advantage of the proposed deep learning approach is in its ability to provide the top factors contributing to the risk score for every point in time for each patient (i.e., local interpretability). The distinction between the global and local notions of interpretability (i.e., what features are contributing to the sepsis risk score for the cohort at large versus an individual patient's hourly prediction window) is most notable when dealing with models capable of capturing higher order interactions and temporal trends in the data. As a result, the degree of risk associated with a factor (e.g., temperature) is a function of other factors in a multiplicative sense (e.g., hypothermia and old age are together a greater risk factor than either by itself). Similarly, the temporal context of a risk factor can alter its contribution to a given risk score calculation (e.g., leukocytosis immediately after surgery may not be unexpected and may contribute differently to the risk for sepsis). These multiplicative and temporal factors (which are captured by the DeepAISE model) result in variations in the importance of risk factors when viewed from a local, hourly prediction perspective for each patient. Note that traditional logistic regression models and decision trees are not capable of making such inferences unless the relevant features are hand-crafted by the experts and included in the model.

The DeepAISE algorithm was implemented in a real-time setting and a decision support user interface (UI) was designed to communicate the risk scores to the clinical team. A major barrier to wide adoption of modern machine learning based CDS tools in clinical practice has been their "black box" nature *(30)*. While it is important to design deep learning models with high performance, it is imperative to build models that provide interpretable data to bedside clinicians that can augment their understanding of the disease process and can contribute to the selection and initiation of appropriate treatments. DeepAISE was designed to be transparent by: 1) continually revealing the top causes contributing to the sepsis risk score (see Fig. 2), 2) providing a lower dimensional view of the patients' trajectory (see Fig. 3), and 3) providing a prioritized list of patients at risk for sepsis (see Fig. 6). These three attributes allow the bedside clinician to identify pathologic deviations from expected physiology early and in real-time throughout the duration of patients' hospital admission. Moreover, we have shown that the top causes can be broken down into two categories of positively and negatively contributing factors to the risk score (see Appendix F of supplementary material). Notably, this analysis shed insight on the input features contributing significantly to the sensitivity (positive contributors) and specificity (negative contributors) of DeepAISE (see Fig. 4). Since one of the key limitations of using EHR data is the intermittent nature of laboratory measurements, we hypothesize that one may use the knowledge of the top contributing factors to protocolize the ordering of laboratory tests, to ensure specific updated measurements of these factors are available to the algorithm, thus improving model sensitivity and specificity.

In recent years, several machine learning-based models for early prediction of sepsis and septic shock have been proposed; although variations in experimental design and definitions of sepsis make a direct comparison of these methods impractical. Desautels et al. *(11)* proposed a proprietary machine learning model called InSight to predict sepsis in ICU patients. Their model used a combination of vital signs, pulse oximetry, GCS, and age as input features. An earlier version of this algorithm relied on the Sepsis-3 definition (specifically $t_{SOFA}$) to train its model and was able to reliably identify (detect) patients at the time they had met the Sepsis-3 criteria, with a 4-hours ahead prediction AUC of 0.74, which is comparable to performance reported by Amland et al. *(35)*. Following the Sepsis-3 definition, Nemati et al. *(16)* achieved an AUC of 0.85 for 4 hours ahead prediction of sepsis, by combining 65 data points including low-resolution data from the EHR and high-resolution vital sign time series features from the bedside monitors. The superior performance of DeepAISE in comparison to the abovementioned models can be attributed to employment of an RNN-based model that captures patients' clinical trajectory.

Experimental design can have a pronounced effect on the reported AUC of machine learning algorithms. A commonly utilized method known as the 'case-control' design (which includes the majority of studies involving biomarkers) significantly overestimates the true prevalence of positive labels and can result in highly optimistic reported performances in the literature *(29)* when compared to a 'sequential prediction' design *(16)*. Assuming a sepsis prevalence of 8% in the ICU population (after excluding all cases of sepsis developed before ICU admission), a median time-to-sepsis of 23 hours, and a 6-hours ahead prediction window, typically only 1-2% of the observed windows include a positive label for sepsis. The case-control study design assumes the timing of sepsis is known *a priori* and seeks to show that certain physiological or biomarker signatures preceding this time are significantly different than that of the non-septic control patients. The resulting algorithms, which are tuned to a 50% prevalence of positive labels, tend to produce high rates of false alarms when deployed prospectively.

In general, statistical evaluation methods (such as the AUC) have a limited applicability when evaluating the clinical utility of such algorithms, although they can provide quantitative metrics for the comparison of various algorithms. In practice, performance metrics are only meaningful when coupled with appropriate clinical protocols that describe the course of action in response to the associated risk alerts. Simple clinical actions (such as 'snoozing' the alarm for X hours if the patient did not meet the clinical threshold to initiate therapy) can significantly alter the false-alarm rate (defined as 1-Specificity) and the associated AUC of an algorithm in practice.

While we have strictly adhered to the Sepsis-3 criteria for defining septic labels in our study, it has been noted that this criterion is too stringent and the sensitivity of early detection is lost to an increased specificity *(36, 37)*. The Sepsis-3 criterion utilized in this study is an acausal clinical construct for demarcating the onset time of sepsis, and as such cannot be used in a clinical setting for early detection of sepsis; however, a predictive analytic risk score when trained to predict the associated onset-time can combine the specificity advantages of Sepsis-3 with the benefits of early recognition. Moreover, it is critical to appreciate that making a clinical diagnosis of sepsis carries much greater value than simply identifying 'poor health' or general decompensation. Clinical outcomes have repeatedly been positively impacted by the rapid administration of broad-spectrum antibiotics, IV fluids, and vasopressors if indicated *(6, 9)*. The predictive ability of DeepAISE is remarkable because it means at a minimum that preparatory measures to implement sepsis bundles can begin much earlier and when appropriate, interventions like antibiotic therapy can be initiated sooner. In both instances this has an enormous potential to reduce the expected morbidity and mortality for this condition. Sepsis survivors often suffer from high rates of readmission *(38)* and many survivors of sepsis face life-long, debilitating sequelae as a result of the disease *(39)*. Future extensions of this work will involve assessing the long-term benefits of competing treatment policies using the ATEE framework. Prospective clinical trials will be necessary to confirm this supposition; however, our findings provide significant clinical evidence for a radical change to the sepsis treatment paradigm that has real-time high-dimensional data analysis and model transparency at its center.

**Materials and Methods**

*Study design*

We performed a retrospective study of all patients admitted to the ICUs at two hospitals within the Emory Healthcare system in Atlanta, Georgia from 2014 to 2018. This investigation was conducted according to Emory University Institutional Review Board approved protocol #33069. External validation of the DeepAISE model was performed on the Medical Information Mart for Intensive Care-III (MIMIC-III) ICU database *(28)*, which is a publicly available database containing de-identified clinical data of patients admitted to the Beth Israel Deaconess Medical Center in Boston, Massachusetts from June 2001 to October 2012. Patients 18 years or older were followed throughout their ICU stay until discharge or development of sepsis, according to Sepsis-3 guidelines (1). The labels of $t_{suspicion}$, $t_{SOFA}$, $t_{sepsis-3}$, $t_{eSOFA}$, and $t_{sepsis-CDC}$ are used extensively throughout the work to define key time points and are clearly described in Table 1. For the purpose of defining sepsis, the order-time of antibiotics and cultures were obtained. For the purpose of checking treatment durations, the administration time-stamps of antibiotics was considered.

For the Emory cohort, data from the EHR (Cerner, Kansas City, MO) were extracted through a clinical data warehouse (MicroStrategy, Tysons Corner, VA). High-resolution heart rate and Mean arterial pressure time series at 2 seconds resolution were collected from select ICUs, through the BedMaster system (Excel Medical Electronics, Jupiter, FL), which is a third-party software connected to the hospital's General Electric monitors for the purpose of electronic data extraction and storage of high-resolution waveforms. Patients were excluded if they developed sepsis within the first 4 hours of ICU admission (by analyzing pre-ICU IV antibiotic administration and culture acquisition) or if their length of ICU stay was less than 8 hours or more than 20 days.

The Emory cohort contained a total of 25,820 patients, 1,445 of whom met the Sepsis-3 criterion four hours or later after ICU admission. Out of the 25,820 patients, 70% of them were used for developing the model (training set), 10% were used for hyper-parameter optimization, and the remaining 20% formed the testing set (see Table S6 for a description of the various holdout datasets that have been used for analysis in this paper). The Emory training set contained a total of 18,074 patients out of which 1,003 patients met the Sepsis-3 criterion, and the Emory testing set contained a total of 5,165 patients out of which 287 patients met the Sepsis-3 criterion during their stay in the ICU. The MIMIC-III external validation cohort was split in a similar fashion, and more details can be found in the Appendix E of Supplementary Material. The DeepAISE model was trained and evaluated on both the Emory cohort and the MIMIC-III external validation cohort separately.

*Attributable treatment effect estimates (ATEE)*

We first defined a sepsis treatment policy ($\pi_l^d$) associated with a criterion $d$, $d \in D = \{d_1, d_2, \ldots, d_k\}$, as initiation of antibiotics within $t$ hours of meeting the criterion (e.g., $t_{sepsis-3}$ or $t_{sepsis-CDC}$), with $l \in L = \{l_1, l_2, \ldots, l_m\}$, taking either positive or negative values (we take $t$ to belong to an interval $l$). We considered four different antibiotics administration intervals, namely [-24, –6), [-6, 0), [0, 3), and [3, 24] hours, corresponding to $l_1, \ldots, l_4$, respectively (See Fig. S10). For the purpose of ATEE analysis, these intervals at the corresponding lags or delays in antibiotics administration were considered as treatment levels.

The propensity score was first introduced as a statistical tool in observational studies to select a control cohort for treatment effect estimation for binary treatments and binary outcomes *(40)*. The method of Generalized Propensity Score (GPS) was developed to extend effect estimation to the case of multi-level (or categorical) treatments. More formally, the GPS is the conditional probability that the *n*-th individual with covariates $C^{(n)}$ receives a treatment level, where $C^{(n)}$ corresponds to the n-th patient's covariates measured prior to administration of antibiotics.

The propensity to receive a given level of treatment *l* (e.g. antibiotics within three hours of $t_{sepsis-3}$), denoted by $GPS(l|C^{(n)})$, was then modeled using a *multi-class* feedforward neural network (namely the GPS-network, see Fig. S11(a)), where given the set of covariates $C^{(n)}$ prior to administration of antibiotics, the model returned the patients' propensity to receive a treatment level, *l*. This propensity score enabled us to assess the effects of a given treatment level on the patient outcome (*Y*), through causal inference with discrete treatment levels and binary outcomes *(41, 42)*. Creating an averaged dose response function (ADRF) by mapping the received treatment levels and the associated GPS to the outcome of interest (e.g., hospital survival) allowed us to explore counterfactual reasoning for timing of antibiotics. That is, 'how might a patient's outcome change if the patient was given levels of treatment different from the hospital policy?'. The ADRF function was modeled through another feedforward neural network (namely the survival network, see Fig. S11(b)), followed by isotonic least squares regression to ensure calibrated estimates of outcomes. The input to the ADRF network was a combination of the one-hot encoded vector of the treatment level, and the GPS corresponding to the treatment level.

Given a set of clinical observations and a counterfactual level of treatment, the GPS-network was first used to map the observations to GPS scores for the actual and counterfactual treatment levels. Next, the GPS scores and the actual and counterfactual treatment levels were separately fed into the ADRF model to estimate the expected effect size ($\Delta(\pi_l^d)$) associated with a given policy and treatment level:

$$\Delta(\pi_l^d) = E_n[Y|l, GPS(l|C^{(n)})] - E_n\left[Y|l_{ac}^{(n)}, GPS\left(l_{ac}^{(n)}\Big|C^{(n)}\right)\right] \quad (1)$$

where the first term inside the parentheses on the right-hand-side of the question was the propensity adjusted expected survival associated with the counterfactual treatment level $l$, and the second term was the propensity adjusted expected survival associated with the actual treatment level, $l_{ac}^{(n)}$ for patient $n$. For each criterion and treatment level, we trained the GPS and ADRF networks 50 times, using random initialization of network parameters, to obtain confidence intervals for our estimation of Eq. 1.

*Development of the DeepAISE model*

DeepAISE began producing scores four hours after ICU admission, and it was designed to predict (on an hourly basis) the probability of onset of sepsis within the next 2, 4, 6, 8, 10 and 12 hours. The two distinct characteristics of the model were *a)* utilization of a class of deep learning algorithms for multivariate time series data known as the Gated Recurrent Unit (GRU) that allows for modeling the clinical trajectory of a patient over time *(38)*, and *b)* deployment of a parametric survival model called the Weibull Cox proportional hazards (WCPH), which casts the problem of sepsis prediction to a time-to-event prediction framework and allows for handling of right censored outcomes *(16),* using the features learned from the underlying GRU model. The parametric survival model allows for efficient end-to-end learning of the GRU and the WCPH parameters using standard deep learning optimization techniques *(43)*.

In preparation for time series modeling, the longitudinal data of patients were binned into consecutive windows of 1-hour duration, with the survival data for each of the time bins comprising of three elements: a set of input features *x*, the time to sepsis event $\tau$ and sepsis event indicator *e*. If a sepsis event occurred within the prediction horizon, the time interval $\tau$ would correspond to the duration of time between the time at which sepsis event occurred and the time of collection of features *x*, with the sepsis event indicator *e* set to 1. If sepsis did not occur within the prediction horizon, the time interval $\tau$ would correspond to one hour more than the duration of the prediction horizon, with the sepsis event indicator *e* set to 0 (i.e., a right-censored event). Instead of feeding the features of a patient directly to a WCPH model, we first fed the longitudinal data of patients into a GRU model followed by a feedforward neural network (FFNN) to learn a representation of patient's trajectory at the current time-step, which was then fed to the WCPH model for time-to-event analysis. The WCPH model, a parametric counterpart to the familiar Cox proportional hazards model *(44)*, represents the instantaneous risk of sepsis $H(t)$ as a product of a baseline hazard function, $H_0(t) = \left(\frac{v}{\lambda}\right)\left(\frac{t}{\lambda}\right)^{v-1}$, and the patient-specific sepsis risk which depends on the output of the GRU-FFNN module (or representations of a patient's clinical trajectory at time t):

$$H(t) = H_0(t)\, exp\left(\beta^T f(x_t)\right) \qquad (2)$$

In the above equation, $f(x_t)$ denotes the output from the GRU-FFNN module, $\beta$ represents the patient-specific hazard parameters, $\lambda > 0$ was a scale parameter and $v > 0$ was a shape parameter for the Weibull distribution. The DeepAISE model (schematic diagram shown in Fig. S2) employed a combination of 2-layer stacked GRU-FFNN framework and a modified WCPH model to predict the onset of sepsis at a regular interval of 1 hour. The parameters of the DeepAISE model, including the WCPH weight vector $\beta$, parameters of the GRU-FFNN were then learnt end-to-end by employing a mini-batch stochastic gradient descent approach to minimize the negative-log likelihood of the data. More technical details of the DeepAISE model and learning the model parameters can be found in Appendix B of Supplementary Materials. To make predictions on new patients, we used the trained model parameters and passed the longitudinal patient data sequentially through the GRU-FFNN, and the WCPH model to compute the survival scores, which provided the probability of not getting sepsis over the prediction horizon. The predicted probability of the sepsis event occurring within the prediction horizon was then defined as one minus the survival score.

*Model evaluation and statistical analysis*

For all continuous variables, we have reported median ([25th - 75th percentile]) and used a two-sided Wilcoxon rank-sum test when comparing two populations. For binary variables, we have reported percentages and use a two-sided chi-square test to assess differences in proportions between two populations. The area under receiver operating characteristic (AUC) curves statistics, specificity (1-false alarm rate) and accuracy at a fixed 85% sensitivity level were calculated to measure the performance of the models. We have reported the DeepAISE performance results for four hours ahead prediction for both the training set and testing set on the Emory cohort and the MIMIC-III external validation cohort. Additionally, we have also reported the performance results for 2, 4, 6, 8, 10 and 12 hours ahead prediction of onset of sepsis. Statistical comparison of all AUC curves was performed using the method of DeLong et al. *(45)*.

*Parameter Optimization*

We trained each model for a total of 200 epochs using the Adam optimizer *(43)*, with a learning rate fixed at 1e-2. The mini-batch size was fixed at a total of 1,000 patients (90% control patients, 10% septic patients), with data randomly sampled (with replacement) in every epoch. To minimize overfitting and to improve the generalizability of the model, L1-L2 regularization was used with L2 regularization parameter set to 1e-3 and L1 regularization parameter set to 1e-5. Our final model had 2 GRU layers stacked on top of each other with the size of hidden state being 100 per layer, followed by 1 fully connected layer with the size of the hidden state being 100, and the output of which was fed into a modified Weibull-Cox proportional hazards model for prediction of onset of sepsis. All of the hyper-parameters of the model: Number of GRU layers, the size of hidden state in each of GRU layers, Number of fully connected layers, the size of hidden state in each of the fully connected layers, learning rate, mini-batch size, L1 regularization parameter, and L2 regularization parameter were optimized using Bayesian optimization *(46)*. All pre-processing of the data was performed using Numpy *(47)*, with the rest of the pipeline implemented using TensorFlow *(48)*.

*Testing the validity of the relevance scores and model interpretability*

For a given risk score calculation the contribution of the individual input features was calculated using the associated relevance scores, by calculating the gradient of the sepsis risk score with respect to all input features and element-wise multiplication by the corresponding input features. The resultant scores, also known as the *relevance scores*, were then z-scored and the top 10 features (i.e., most frequently observed features across patients and across time) with a z-score of larger than 1.96 (corresponding to a 95% confidence interval) were reported for analysis of the overall importance of input risk factors (or 'global interpretability'). To test the hypothesis that the individual relevance scores capture meaningful information about the contribution of each input feature to the risk scores, we performed a series of studies in which we systematically masked (or treated as missing data) 1) the top 10 features with the largest positive and negative relevance scores in a global sense (*global feature replacement analysis*), 2) the top 10 locally important features for each individual risk scores (*local feature replacement analysis*) and 3) a random set of 10 features at each point in time (repeated 100 times), and calculated the resulting risk scores produced by DeepAISE and the corresponding AUCs. This analysis allowed us to systematically compare the contribution of the locally important features against the globally important predictors and against the 95-percentile of a randomly selected set of features.

*Real-time implementation details*

To facilitate real-time decision support with DeepAISE, a software platform was developed. The platform is responsible for interfacing with an EHR to obtain real-time patient data, computing hourly DeepAISE scores, and presenting the results in an interpretable, clinically meaningful fashion. The platform consists of four distinct microservices as shown in Fig. 5. These microservices are containerized using Docker. The four microservices are:

a) **DeepAISE Model Service**: The DeepAISE model is wrapped in a python wrapper and exposes a REST API that allows an external entity to run the model, in a stateless inference mode, and return a DeepAISE score given a set of observations about a patient. The stateless nature refers to the fact that, in a clustered deployment of the model, predictions about a patient are not tied to a specific worker node. This makes the system highly scalable and it can leverage popular orchestration systems such as Kubernetes and Docker Swarm to scale on-demand, in response to fluctuations in system load. Finally, for security needs, the system maintains an audit log of every request it receives, and whether the request was successfully executed.

b) **Results Database Service**: This is the core data management layer of the DeepAISE platform and consists of MongoDB and a Java based web service that is responsible for providing a REST API as well as the necessary security and auditing infrastructure. The data is organized as binary JSON documents, where each document represents the raw information (observations, measurements, etc.) and the DeepAISE predictions (scores, contributing factors etc.) about a specific patient, at a specific point in time. The organization of the data in this fashion allows us to make time series requests and query for population trends in the past $n$ hours as well as take deeper dives into an individual patient's data.

c) **Data Orchestrator Service**: This is the data system that interfaces with the healthcare IT systems. It is responsible for fetching patient data (observations, measurements etc.), and transform them into the JSON document structure described above. It posts this data to the DeepAISE Model Service and then incorporates the returned results into the JSON document before posting them to the Results Database Service. It also maintains systems information, tracking uptime, as well as population characteristics that could be used to detect anomalies in data

d) **User Interface Service**: This is the service that serves the user facing interfaces. It presents a dashboard that provides a ranked presentation of patients at risk for sepsis (see Fig.6, 'Under Observation' column). Creating a card for each patient that concisely displayed their sepsis risk on the front and provided the top contributing factors on the card's opposite side made it easier for the clinical team to investigate impending cases of sepsis. A second drag-and-drop column allowed for clinicians to identify patients that have been reviewed and for whom no immediate action was deemed necessary (see Fig.6, 'Snoozed Alarms' column). Situational awareness was further improved by the movement of septic patients to a third column (see Fig.6, 'Treatment Initiated' column) indicating that a sepsis related treatment had been initiated.

A web-based version of the DeepAISE software pipeline, with anonymized data, is currently available in demo-mode (https://sepsis.app/ ; username:demo, password:demo).

**Acknowledgments:** We would like to acknowledge our funding sources. Additionally, we would like to thank Drs. Andre Holder and Russell Jeter for their constructive comments on the manuscript, and the members of our tele-ICU team especially Dr. Timothy G. Buchman and Cheryl Hiddleson for feedback regarding the user interface, Chad Robichaux for help with data extraction and curation from the Emory Clinical Data Warehouse, and Fatemeh Amrollahi for her contribution to development of the software pipeline. **Funding:** Dr. Nemati is funded by the National Institutes of Health (NIH), award #K01ES025445. Dr. Josef is funded by the Surgical

Critical Care Initiative (SC2i), funded by the Department of Defense Defense Health Program Joint Program Committee 6 / Combat Casualty Care (USUHS HT9404-13-1-0032 and HU0001-15-2-0001). The opinions or assertions contained herein are the private ones of the author and are not to be construed as official or reflecting the views of the Department of Defense, the Uniformed Services University of the Health Sciences, the NIH or any other agency of the US Government.

**Author contributions:** S.P.S. and S.N. conceived the overall study, developed the network architectures, conducted the experiments, and analyzed the data. C.J. provided clinical expertise, reviewed patient data and contributed to interpretation of results and the write-up. S.P.S., S.N. and A.S. contributed to the software engineering. S.P.S. and C.J. prepared all the figures. S.P.S. wrote the initial draft of the manuscript. S.P.S., C.J., A.S. and S.N. wrote and edited the final manuscript.

**Competing interests:** The authors declare that they have no competing interests. **Data and materials availability:** MIMIC-III is a de-identified critical care dataset and is publicly available. Access to a de-identified data has been made available as a part of the PhysioNet Challenge 2019. Access to the computer code used in this research is available upon request to the corresponding author.

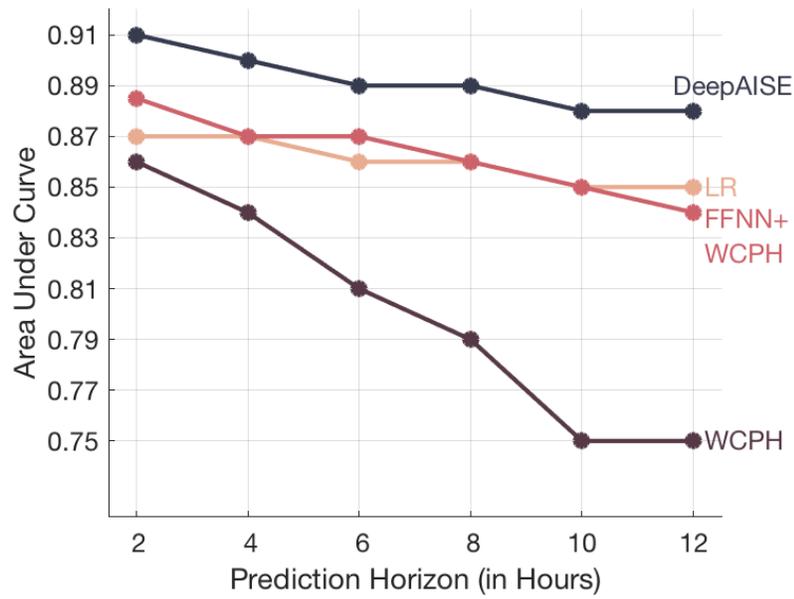

**Fig. 1: Comparison of DeepAISE performance on Emory testing set for prediction horizons of 2, 4, 6, 8, 10 and 12 hours.** Comparison of DeepAISE performance with other baseline models.

*DeepAISE = Deep Artificial Intelligence Sepsis Expert*

*LR = Logistic Regression*

*FFNN = Feedforward Neural Network*

*WCPH = Weibull Cox Proportional Hazard layer*

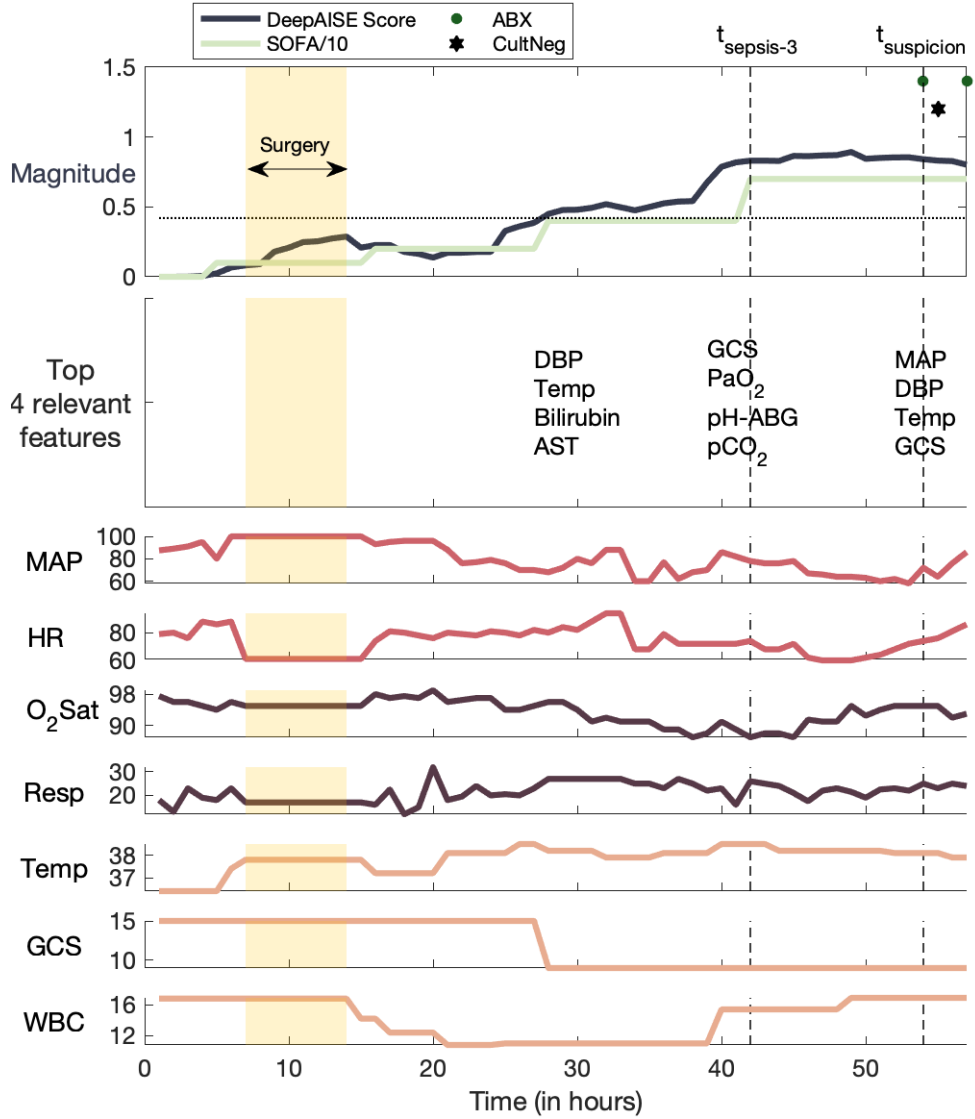

**Fig. 2: A clinically interpretable example of DeepAISE.** The DeepAISE score is shown for a septic patient according to the Third International Consensus Definitions for Sepsis (Sepsis-3). Commonly recorded hourly vital signs of the patient, including heart rate (HR), mean arterial blood pressure (MAP), respiratory rate (RESP), temperature (TEMP), oxygen saturation ($O_2$Sat) are shown. The most significant features contributing to the DeepAISE score are listed immediately below the DeepAISE Scores (for clarity of presentation, only selected time points are shown). The horizontal dashed line indicates the prediction threshold corresponding to a sensitivity of 0.85. Refer to Appendix C of Supplementary Material for more details on the abbreviated features.

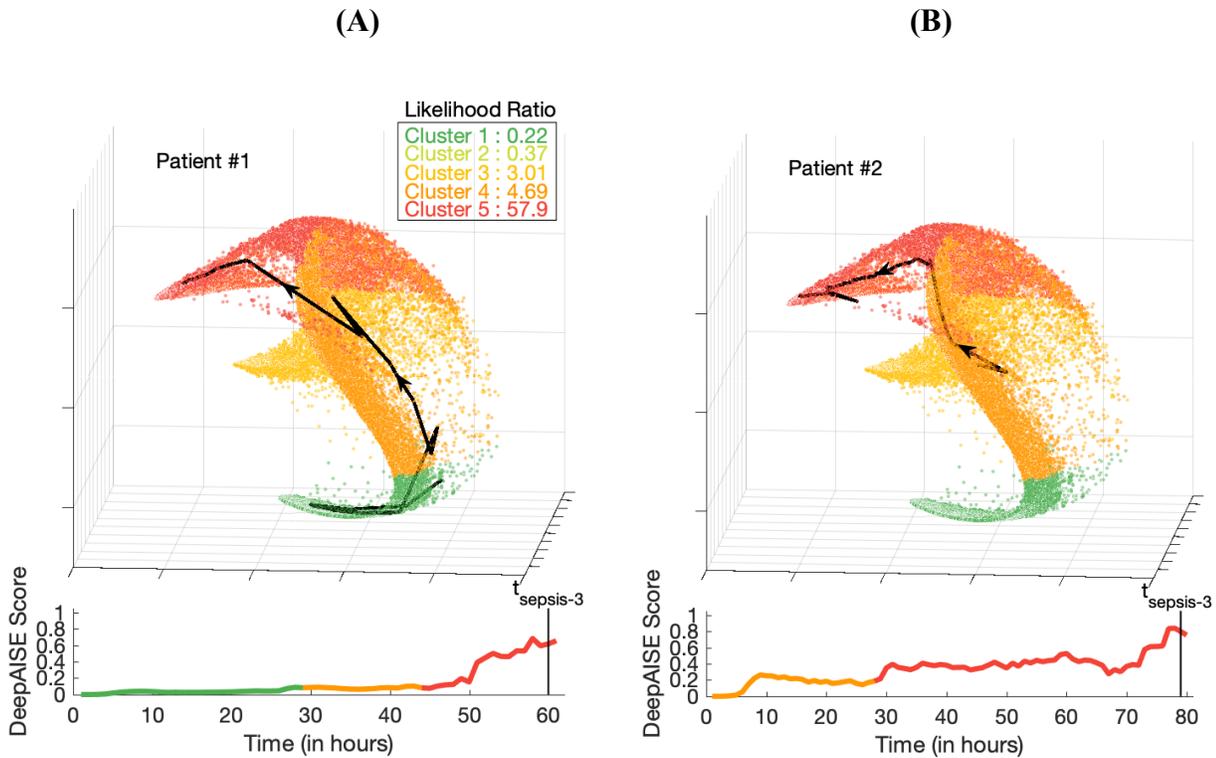

**Fig. 3: Visualization of DeepAISE time series covariates performed by spectral clustering.** The linear trajectory of the DeepAISE score for 2 septic patients from ICU admission until sepsis diagnosis is displayed below a larger manifold that makes use of spectral clustering to visually display a patient's physiologic journey through their illness (each point on the graph represents one hour of data from one patient). The colors for both patients in the plots were chosen based on the predicted sepsis risk score (green represents the lowest predicted sepsis risk score, and red represents the highest predicted sepsis risk score). A similar figure with septic patients highlighted is shown in Fig. S5 of Supplementary Material. **(A)** Patient #1 (P1) was a 63-year-old female admitted for a left sided subdural hemorrhage who underwent a craniectomy on hospital day zero. This patient remained intubated after surgery and began receiving treatment

for a culture proven ventilator associated pneumonia the afternoon of hospital day number three. DeepAISE identified this patient as being septic nearly 24hrs before clinical treatment was implemented (See Fig. S6). **(B)** Patient #2 (P2) was a 70 year old male who was admitted for altered mental status and seizures after vascular coiling of a middle cerebral artery (MCA) aneurysm. P2 was intubated on admission and began treatment for a culture proven ventilator associated pneumonia on hospital day five however DeepAISE made its sepsis prediction nearly 36 hours prior to this time, after demonstrating a steadily worsening score since admission (See Fig. S7).

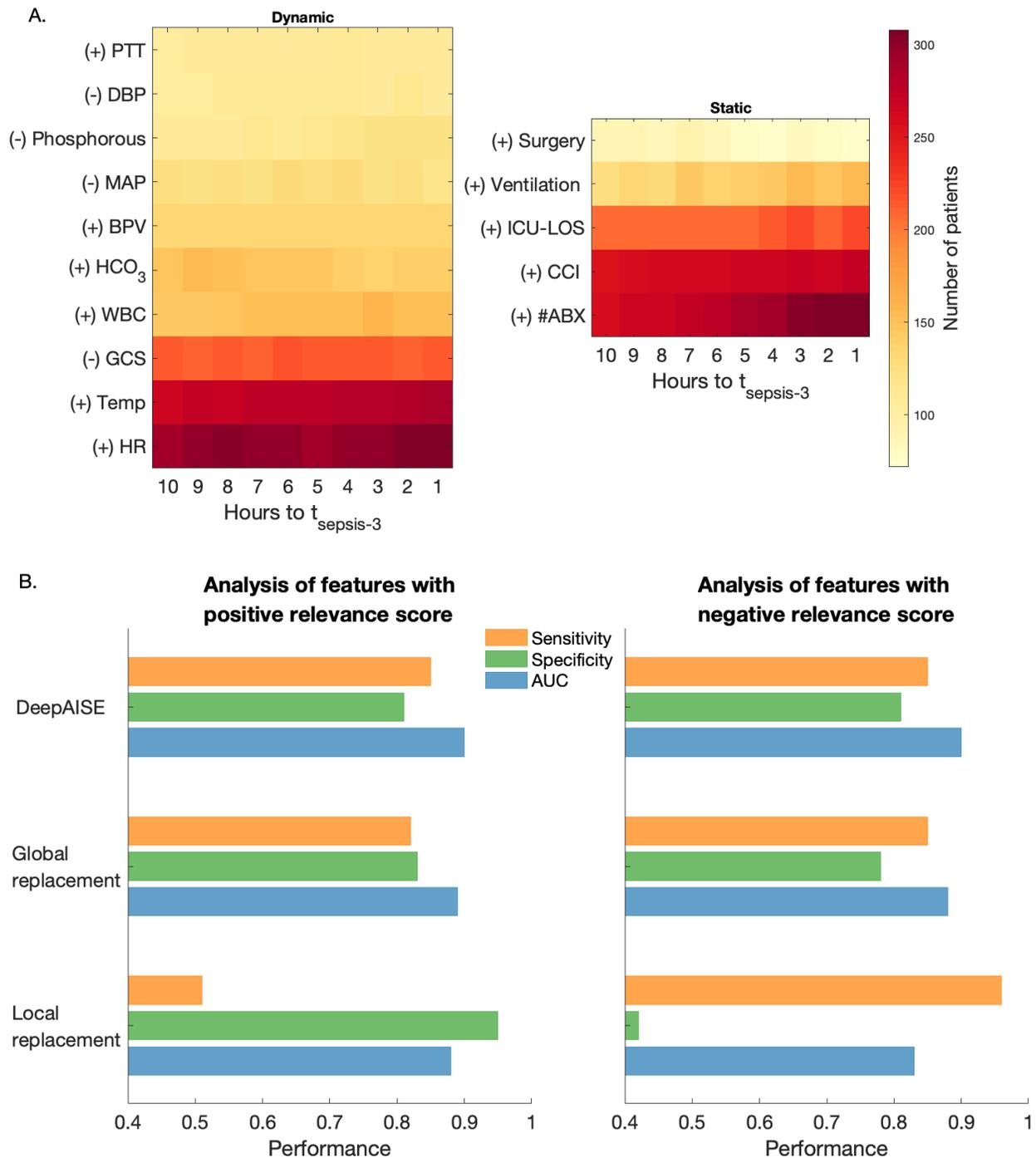

**Fig. 4: Most common features contributing to an elevated risk score. (A)** Every hour DeepAISE identifies the top features contributing to an individual septic patient's risk score. The left subfigure demonstrates the frequency of the top ten dynamic features (ordered according to the magnitude of the relevance score) across the septic patient population (in the Emory cohort) preceding $t_{sepsis-3}$ and the right subfigure demonstrates the frequency of the top five static features that are seen preceding $t_{sepsis-3}$. Features with positive gradient with respect to the sepsis risk score are identified by '(+)'. Features with negative gradient with respect to the sepsis risk score are identified by '(-)'. **(B)** Summary of performance of DeepAISE (on the Emory testing set) when *global feature replacement analysis* and *local feature replacement analysis* were performed for features with positive relevance score (left subfigure; see Table S4) and negative relavance score (right subfigure; see Table S5). Note that the performance (AUC) of DeepAISE when a random set of 10 features at each point in time were masked (repeated 100 times) was 0.899 [0.886, 0.901]. The sensitivity and specificity values reported for *global feature replacement analysis* and *local feature replacement* analysis were measured at threshold corresponding to 0.85 sensitivity for the original model.

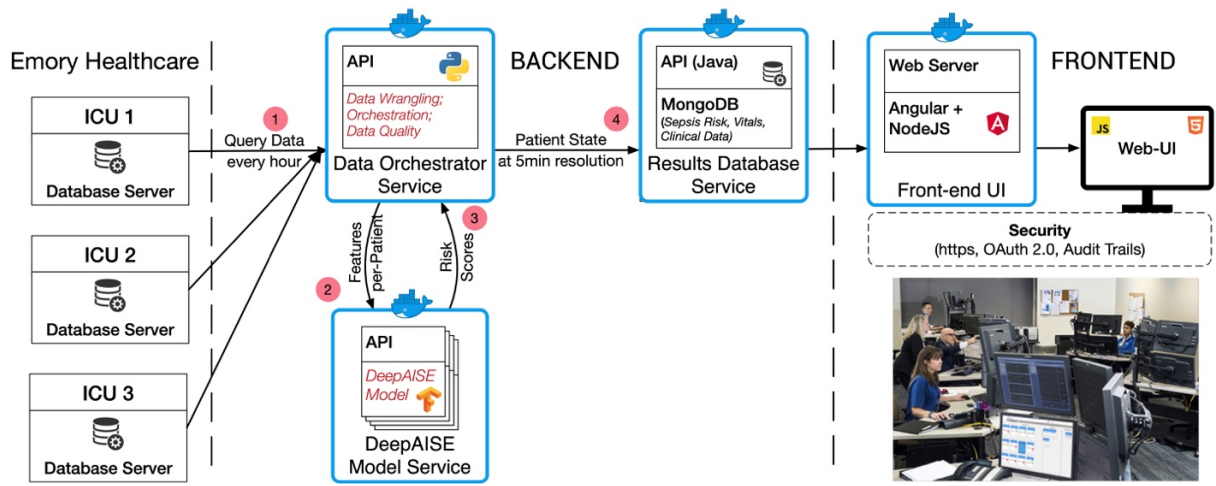

**Fig. 5: Block diagram of the DeepAISE software platform.** EHR is queried for the required data elements (1). EHR data is then prepared by the Data Orchestrator Service and passed to the DeepAISE Model Service for DeepAISE score computation (2). The resulting scores are computed and returned to the Data Orchestrator Service using a predefined set of API calls (3). The scores are then managed by a time series database in the Results Database Service (4), which provides the required data for the UI implemented using client-side Javascript. The figure inset shows a live integration of DeepAISE UI in a tele-ICU workflow.

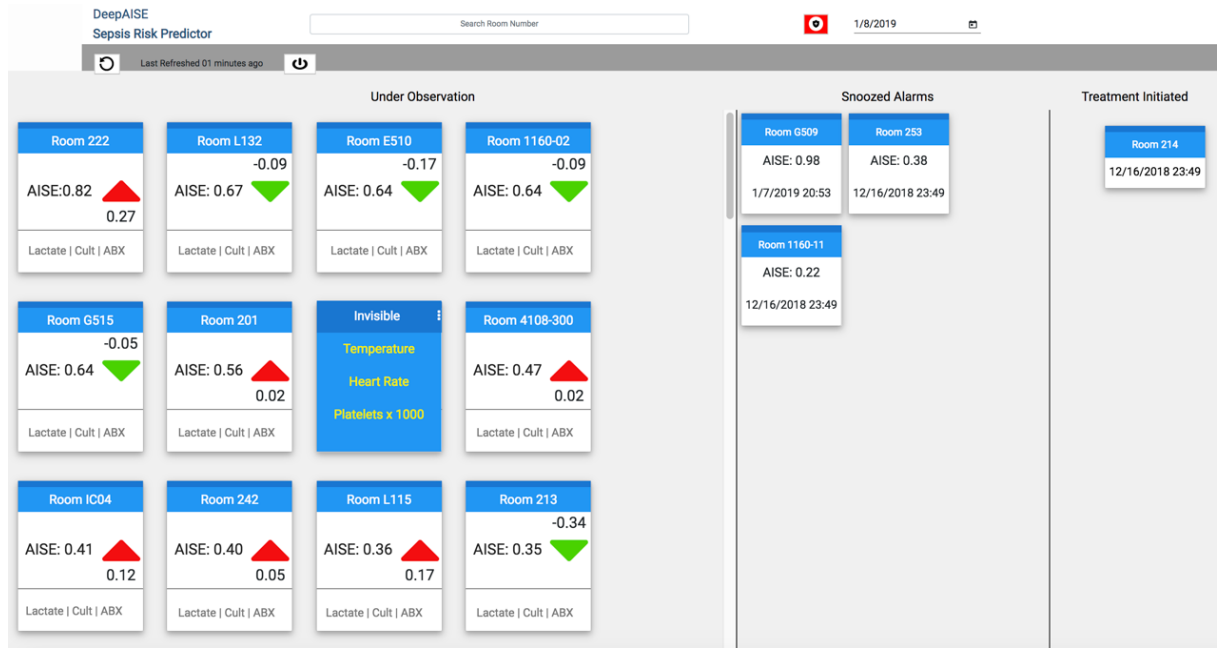

**Fig. 6: Screenshot of the clinician facing DeepAISE UI.** In the left column, patients are ranked in decreasing severity of illness. An individual patient card shows DeepAISE score on the front, and upon a single mouse click the card is turned displaying the top causes contributing to the risk score (e.g. Temperature, Heart Rate, Platelets). The middle column displays patients that have undergone review by a clinician. The right most column displays patients for whom treatment has been initiated.

Table 1: Description of defined time points utilized in the study.

| Time Point | Criteria |
|---|---|
| $t_{suspicion}$ | Clinical suspicion of infection identified as the earlier timestamp of antibiotics and blood cultures within a specified duration. (If antibiotics were given first, the cultures must have been obtained within 24 hours. If cultures were obtained first, then antibiotic must have been subsequently ordered within 72 hours). |
| $t_{SOFA}$ | The occurrence of end organ damage as identified by a two-point deterioration in SOFA score within a 6-hour period |
| $t_{sepsis-3}$ | The onset time of sepsis-3 is marked when both $t_{suspicion}$ and $t_{SOFA}$ have happened within close proximity to each other. Specifically, $t_{SOFA}$ must occur 24 hours before $t_{suspicion}$ or up to 12 hours after the $t_{suspicion}$ ($t_{SOFA}$ + 24 hours > $t_{suspicion}$ > $t_{SOFA}$ - 12 hours). The earlier of the $t_{SOFA}$ or $t_{suspicion}$ was assigned to $t_{sepsis}$. |
| $t_{eSOFA}$ | The occurrence of end organ damage as identified by one point or higher eSOFA score within a 6-hour period |
| $t_{sepsis-CDC}$ | The onset time of sepsis-CDC is marked when both $t_{suspicion}$ and $t_{eSOFA}$ have happened within close proximity to each other. Specifically, $t_{eSOFA}$ must occur 24 hours before $t_{suspicion}$ or up to 12 hours after the $t_{suspicion}$ ($t_{eSOFA}$ + 24 hours > $t_{suspicion}$ > $t_{eSOFA}$ - 12 hours). The earlier of the $t_{eSOFA}$ or $t_{suspicion}$ was assigned to $t_{sepsis-CDC}$. |

Table 2: Summary of percent improvement in survival (median[Interquartile range]) as a function of elapsed time from different sepsis onset times to antibiotic start time using Attributable treatment effect estimates (ATEE).

|  | [-24,-6) hours | [-6,0) hours | [0,3) hours | [3,24] hours |
|---|---|---|---|---|
| $t_{sepsis-3}$, (%) | 8.7 [7.6, 10.2] | 8.2 [7.4, 9.9] | 3.3 [2.1, 4.0] | 2.8 [1.9, 3.5] |
| $t_{sepsis-CDC}$, (%) | 9.7 [7.3, 10.7] | 0.8 [-0.2, 5.3] | -3.0 [-4.1, -1.4] | 1.1 [0.1, 2.3] |

**Table 3: Summary of DeepAISE performance.**

| Performance Metric: *testing set (training set)* | 4 hours | 6 hours | 12 hours |
|---|---|---|---|
| *Emory cohort* | | | |
| AUC | 0.90 (0.94) | 0.89 (0.90) | 0.88 (0.89) |
| Specificity | 0.80 (0.89) | 0.78 (0.84) | 0.73 (0.78) |
| *MIMIC-III validation cohort* | | | |
| AUC | 0.87 (0.90) | 0.86 (0.87) | 0.84 (0.86) |
| Specificity | 0.74 (0.78) | 0.72 (0.75) | 0.69 (0.73) |

\* *AUC = Area under the Curve*

*Sensitivity was fixed at 0.85*